\documentclass{article}

\PassOptionsToPackage{numbers, compress}{natbib}
     \usepackage[preprint]{neurips_2019}

\usepackage[utf8]{inputenc} % allow utf-8 input
\usepackage[T1]{fontenc}    % use 8-bit T1 fonts
\usepackage{hyperref}       % hyperlinks
\usepackage{url}            % simple URL typesetting
\usepackage{booktabs}       % professional-quality tables
\usepackage{amsfonts}       % blackboard math symbols
\usepackage{nicefrac}       % compact symbols for 1/2, etc.
\usepackage{microtype}      % microtypography

\usepackage{graphicx}
\usepackage{caption}
\usepackage{subcaption}
\usepackage{amsmath}

\usepackage{algorithm}
\usepackage[noend]{algpseudocode}

\usepackage{csquotes}

\usepackage[dvipsnames]{xcolor}
\usepackage{listings}
\usepackage{textcomp}
\definecolor[named]{DarkBlue}{cmyk}{1,1,0,0.45}
\definecolor[named]{Sepia}{cmyk}{0,0.411,0.821,0.561}
\definecolor[named]{DarkOrchid}{cmyk}{0.25,0.75,0,0.20}
\definecolor[named]{DarkGreen}{cmyk}{1,0,1,0.61}
\lstdefinestyle{Gen}{
  basicstyle=\ttfamily\scriptsize,
  columns=fullflexible,
  keepspaces=true,
  alsoletter={\.,\%,\#, \@, \?, \/, \!},
  morekeywords=[1]{function, if, else, elseif, end, for, begin, in, const, struct, using, return, const, do, quote},
  keywordstyle=[1]\textcolor{Sepia},
  morekeywords=[2]{\@static, \@gen, \@inj, \@trace, \@param, \@input, \@output,
                  \@tf_function, \@call, \@read, \@tf_call, \@copy,
                  \@splice, \@write,\@choicediff,\@diff,\@ad,\@tensorflow_function},
  keywordstyle=[2]\textcolor{DarkBlue},
  morekeywords=[3]{initialize, generate, choicemap, map_optimize, importance_resampling, select, mala, mh, metropolis_hastings, update, propose, assess, custom_importance, train_batched!, TFFunction},
  keywordstyle=[3]\bfseries,
  morekeywords=[4]{normal,bernoulli,gamma,uniform,beta,independent_pixel_noise},
  keywordstyle=[4]\textcolor{DarkGreen},
  showstringspaces=False,
  morecomment=[l]{\#},
  commentstyle=\color{darkgray}\ttfamily,
  escapeinside={<@}{@>},
  numbers=left,
  numberstyle=\tiny\color{darkgray},
  numbersep=3pt
}

\lstdefinestyle{Assignment}{
  basicstyle=\ttfamily\scriptsize,
  columns=fullflexible,
  keepspaces=true,
  alsoletter={\.,\%,\#, \@, \?, \/},
  morekeywords=[1]{function, if, else, end, for, begin, in, const, struct, using, return, const},
  keywordstyle=[1]\textcolor{Sepia},
  morekeywords=[2]{\@gen, \@inj, \@trace, \@param, \@input, \@output,
                  \@tf_function, \@call, \@read, \@tf_call, \@copy,
                  \@splice, \@write},
  keywordstyle=[2]\textcolor{DarkBlue},
  morekeywords=[3]{generate, choicemap, map_optimize, metropolis_hastings, select},
  keywordstyle=[3]\bfseries,
  morekeywords=[4]{normal,bernoulli},
  keywordstyle=[4]\textcolor{DarkGreen},
  showstringspaces=False,
  morecomment=[l]{\#},
  commentstyle=\color{darkgray}\ttfamily,
  escapeinside={<@}{@>},
  moredelim=**[is][\color{red}]{@@}{@@}
}

\title{Bayesian causal inference via \\probabilistic program synthesis}

\author{
  Sam Witty\thanks{Contributed equally.}\\
  University of Massachusetts\\
  Amherst, MA 01002 \\
  \texttt{switty@cs.umass.edu} \\
   \And
   Alexander Lew\footnotemark[1]\\
   Massachusetts Institute of Technology \\
   Cambridge, MA 02139\\
   \texttt{alexlew@mit.edu} \\
   \AND
  David Jensen \\
  University of Massachusetts\\
  Amherst, MA 01002 \\
  \texttt{jensen@cs.umass.edu} \\
   \And
   Vikash Mansinghka \\
   Massachusetts Institute of Technology \\
   Cambridge, MA 02139\\
   \texttt{vkm@mit.edu} \\
}

\begin{document}

\maketitle
\begin{abstract}
  Causal inference can be formalized as Bayesian inference that combines a prior distribution over causal models and likelihoods that account for both observations and interventions. We show that it is possible to implement this approach using a sufficiently expressive probabilistic programming language. Priors are represented using probabilistic programs that generate source code in a domain specific language. Interventions are represented using probabilistic programs that edit this source code to modify the original generative process. This approach makes it straightforward to incorporate data from atomic interventions, as well as shift interventions, variance-scaling interventions, and other interventions that modify causal structure. This approach also enables the use of general-purpose inference machinery for probabilistic programs to infer probable causal structures and parameters from data. This abstract describes a prototype of this approach in the Gen probabilistic programming language.
\end{abstract}

\section{Introduction}

Bayesian formulations of causal inference enable practitioners to explicitly reason about uncertainty when answering structural questions (e.g., \enquote{What is the probability that $X$ causes $Y$?}) as well as questions about the effects of a specific intervention (e.g., \enquote{How much will intervening to make $X=x'$ increase the probability that $Y=y'$?}). Bayesian formulations have been developed for both the potential outcomes framework~\cite{mccandless2009bayesian} and the causal graphical models framework~\cite{friedman2000being, griffiths2009theory, heckerman1995learning}. In principle, Bayesian approaches make it possible to incorporate prior knowledge and make efficient use of limited data~\cite{mansinghka2006structured, murphy2012machine}.

In this paper, we explore a new approach to implementing Bayesian causal inference based on probabilistic programming, inspired by Bayesian synthesis~\cite{saad2019bayesian}. Probabilistic programming languages enable users to compactly specify probabilistic models in code. Some languages, like Stan \cite{carpenter2017stan}, have syntax that closely resembles the statistical notation often used in the literature to define probabilistic models: a list of equations of the form $x \sim \dots$. Others, like Gen~\cite{cusumano-towner2019gen}, allow users to include arbitrary program control flow in their models; a model is represented by a program that simulates stochastically from a distribution. In this paper, we represent hypothesized causal models explaining some phenomenon as programs in \textit{MiniStan}, a simple probabilistic programming language designed to resemble Stan (Figure~\ref{fig:grammar}). Then, we use a more expressive probabilistic programming language, Gen, to encode a prior and likelihood over MiniStan programs, and to do inference. The Gen model (i) stochastically generates MiniStan programs to encode a prior distribution over causal model structures and parameters, (ii) programmatically edits the generated MiniStan programs to reflect interventions and experimental conditions, then (iii) interprets the MiniStan programs to generate observational and experimental data. We can then use Gen's inference programming and conditioning features to condition the entire process on actual observational and experimental data, and to obtain posterior samples of the MiniStan code defining the original observational model\textemdash that is, to perform both structure learning and parameter estimation.

% This paper shows how to use probabilistic programming to formalize Bayesian inference of the structure and parameters of causal models, incorporating both observational data and multiple types of experimental data. Prior distributions on causal models are represented using probabilistic programs that generate source code in a domain specific language. As in recent work on Bayesian synthesis of probabilistic programs~\cite{saad2019bayesian}, observational data can be incorporated by writing an interpreter for this domain-specific language inside a sufficiently expressive probabilistic programming language. Structure learning and parameter estimation can then be implemented using inference programming constructs.

\begin{figure*}[b!]
\begin{align*}
    P &\rightarrow S \mid S; \, P && \text{Programs}\\
    S &\rightarrow x = E \mid x \sim D && \text{Statements}\\
    D &\rightarrow \texttt{normal}(E, E) \mid \texttt{uniform}(E, E) \mid \texttt{bernoulli}(E) &&\text{Distributions}\\
    E &\in \text{deterministic Julia expressions}\\
    x &\in \text{Julia variable identifiers}
\end{align*}
\caption{Grammar of MiniStan}
\label{fig:grammar}
\end{figure*}

Causal models are typically structured as a set of autonomous components \cite{aldrich1989autonomy,haavelmo1944probability,pearl2000causality}, such that interventions in the system can be accurately represented in the model as an alteration of a small number of model components, and all other model components (and the causal relationships among them) remain unchanged. In the formalism of causal graphical models, interventions are typically expressed using the do-operator \cite{pearl2000causality}, which fixes the value of one random variable and removes the influence of its parents. However, many realistic interventions are not accurately represented by this particular variety of model alteration \cite{eberhardt2007interventions,korb2004varieties,sherman2019intervening}. For example, realistic interventions might best be represented by altering the functional form of a particular dependence, enabling or disabling specific causes, or enacting complex combinations of these interventions. This paper demonstrates interventions represented as modifications of probabilistic program source code and shows how this representation enables the Bayesian synthesis approach to handle a broad class of experimental data.

\subsection{A conceptual example}

\begin{figure*}[t!]
    \centering
    \begin{subfigure}[t]{0.4\textwidth}
        \label{fig:graph_edge}
        \centering
        \includegraphics[height=1.2in]{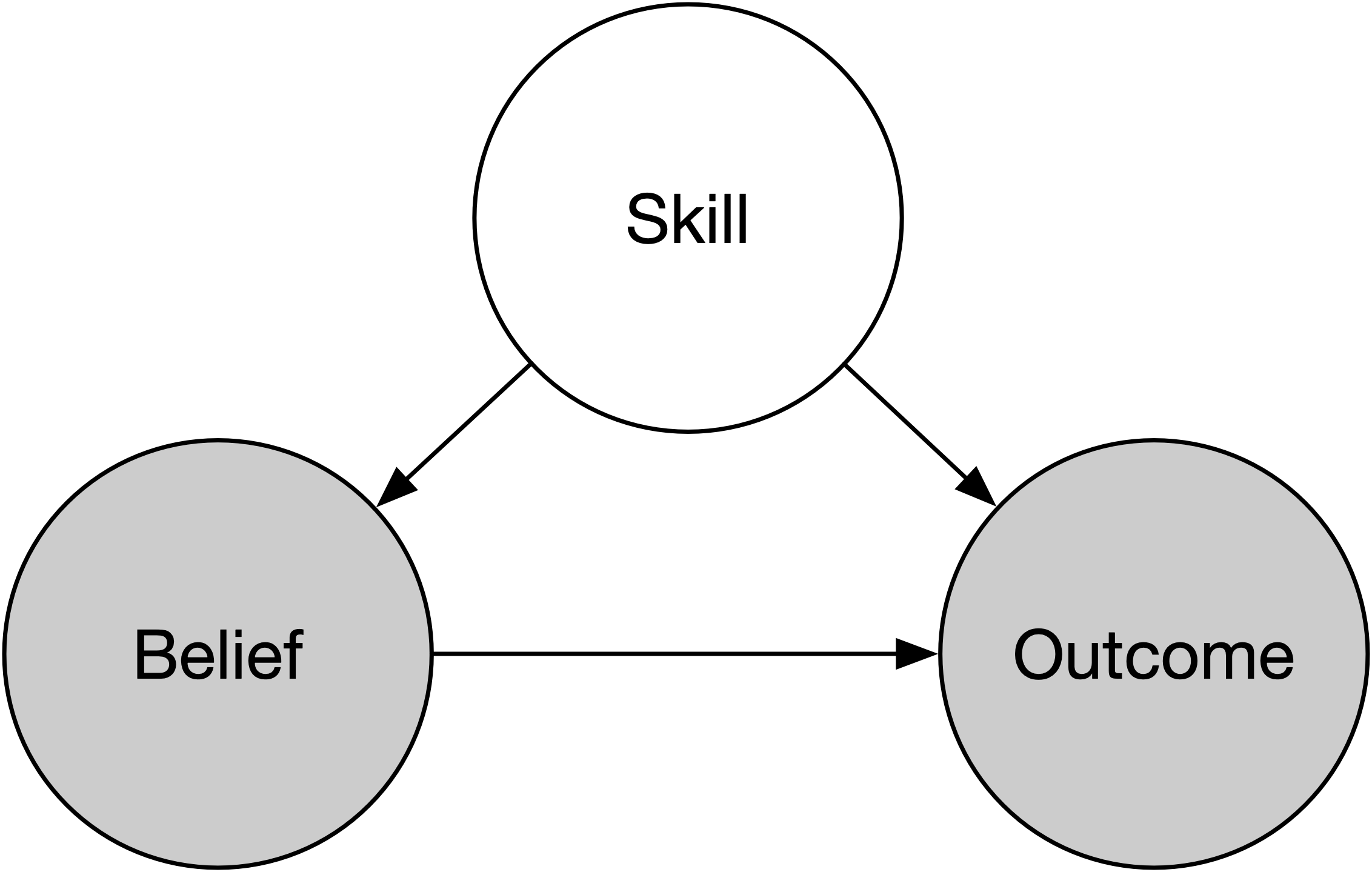}
        \caption{\enquote{Belief and skill matter} CGM.}
    \end{subfigure}
    ~
    \begin{subfigure}[t]{0.4\textwidth}
        \label{fig:graph_no_edge}
        \centering
        \includegraphics[height=1.2in]{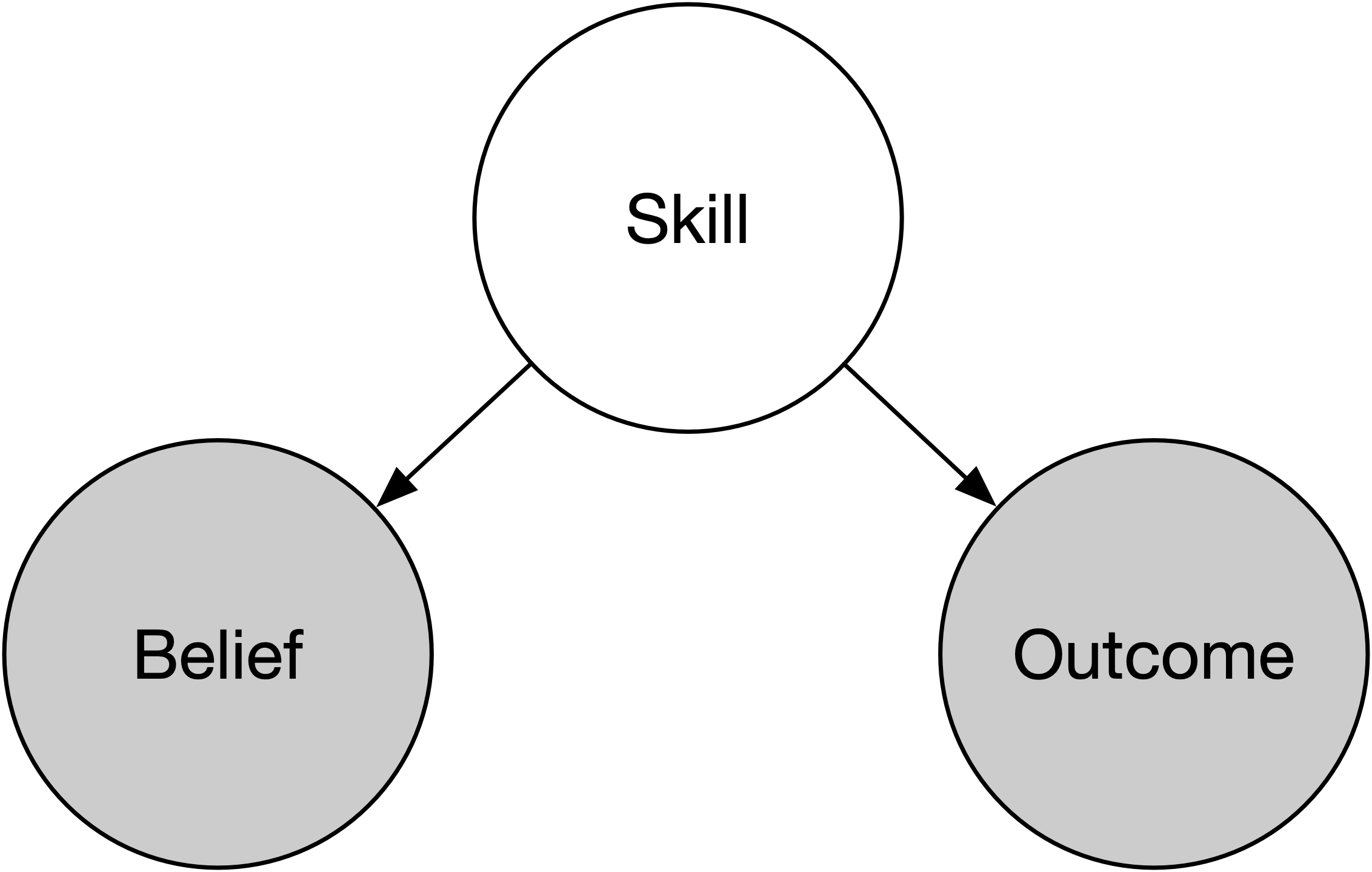}
        \caption{\enquote{Only skill matters} CGM.}
    \end{subfigure}
    
    \begin{subfigure}[t]{0.49\textwidth}
        \label{fig:code_edge}
\begin{lstlisting}[frame=tb, framesep=5pt, xrightmargin=4pt, xleftmargin=0pt, style=Gen, numbers=none]
quote
  s ~ normal(mu_s, sigma_s)
  b ~ normal(s, sigma_b)
  logit_o = s * lambda_so + b * lambda_bo
  o ~ bernoulli(1/(1+exp(-logit_o)))
end
\end{lstlisting}
        \caption{\enquote{Belief and skill matter} model as source code.}
    \end{subfigure}
    ~
    \begin{subfigure}[t]{0.49\textwidth}
        \label{fig:code_no_edge}
\begin{lstlisting}[frame=tb, framesep=5pt, xrightmargin=20pt, xleftmargin=4pt, style=Gen, numbers=none]
quote
  s ~ normal(mu_s, sigma_s)
  b ~ normal(s, sigma_b)
  logit_o = s * lambda_so
  o ~ bernoulli(1/(1+exp(-logit_o)))
end
\end{lstlisting}
        \caption{\enquote{Only skill matters} model as source code.}
    \end{subfigure}
    \caption{A conceptual example combining structure learning and parameter estimation.}
    \label{fig:model_intro}
\end{figure*}

Consider the task of inferring whether a student's belief in her ability is causal for success at a research project. Observational data on student belief and student success alone are insufficient to answer this question, due to the confounding effect of skill (see Figures~\ref{fig:model_intro}a \& \ref{fig:model_intro}b).

We can imagine multiple types of experiments that would enable effective causal inference despite the confounding effect of skill. For example, an advisor could encourage a student, shifting her belief in her ability (but not increasing her skill). An advisor could also administer an assessment on the key skills needed for the project, before the student attempts it, and look at the results. Unfortunately, although this might reveal the true skill level to the advisor, this might also change the student's belief in her own ability to succeed. Hypothetically, one can imagine a miracle pill that modifies one's confidence to a fixed value, without changing anything else. Each of these experiments corresponds to a different modification to the source code from Figures~\ref{fig:model_intro}c and~\ref{fig:model_intro}d. Examples of these modifications are shown in Figures~\ref{fig:interventions}a-f.

This paper shows how to formalize this example, using probabilistic programs that generate, edit, and interpret the source code of causal models. It also presents results from an implementation in the Gen probabilistic programming language, demonstrating the utility of incorporating diverse sources of experimental data.

\begin{figure*}[b!]
    \begin{subfigure}[t]{0.49\textwidth}
        \label{fig:belief_pill_code_edge}
\begin{lstlisting}[frame=tb, framesep=3pt, xrightmargin=4pt, xleftmargin=0pt, style=Gen, numbers=none]
quote
  s ~ normal(mu_s, sigma_s)
  b <@\textcolor{BrickRed}{= 5}@>
  logit_o = s * lambda_so + b * lambda_bo
  o ~ bernoulli(1/(1+exp(-logit_o)))
end
\end{lstlisting}
\vspace{-2mm}

\caption{\enquote{Belief and skill matter} with belief pill.}
    \end{subfigure}
    ~
    \begin{subfigure}[t]{0.49\textwidth}
        \label{fig:belief_pill_code_no_edge}
\begin{lstlisting}[frame=tb, framesep=3pt, xrightmargin=4pt, xleftmargin=0pt, style=Gen, numbers=none]
quote
  s ~ normal(mu_s, sigma_s)
  b <@\textcolor{BrickRed}{= 5}@>
  logit_o = s * lambda_so
  o ~ bernoulli(1/(1+exp(-logit_o)))
end
\end{lstlisting}   
\vspace{-2mm}
\caption{\enquote{Only skill matters} with belief pill.}
    \end{subfigure}
    
    \begin{subfigure}[t]{0.49\textwidth}
        \label{fig:encouragement_code_edge}
\begin{lstlisting}[frame=tb, framesep=3pt, xrightmargin=4pt, xleftmargin=0pt, style=Gen, numbers=none]
quote
  s ~ normal(mu_s, sigma_s)
  b ~ normal(s <@\textcolor{BrickRed}{+ 3}@>, sigma_b)
  logit_o = s * lambda_so + b * lambda_bo
  o ~ bernoulli(1/(1+exp(-logit_o)))
end
\end{lstlisting}
\vspace{-2mm}

        \caption{\enquote{Belief and skill matter} with encouragement design.}
    \end{subfigure}
    ~
    \begin{subfigure}[t]{0.49\textwidth}
        \label{fig:encouragement_no_edge}
\begin{lstlisting}[frame=tb, framesep=3pt, xrightmargin=4pt, xleftmargin=0pt, style=Gen, numbers=none]
quote
  s ~ normal(mu_s, sigma_s)
  b ~ normal(s <@\textcolor{BrickRed}{+ 3}@>, sigma_b)
  logit_o = s * lambda_so
  o ~ bernoulli(1/(1+exp(-logit_o)))
end
\end{lstlisting}
\vspace{-2mm}
        \caption{\enquote{Only skill matters} with encouragement design.}
    \end{subfigure}
    
    \begin{subfigure}[t]{0.49\textwidth}
        \label{fig:assessed_code_edge}
\begin{lstlisting}[frame=tb, framesep=3pt, xrightmargin=4pt, xleftmargin=0pt, style=Gen, numbers=none]
quote
  s ~ normal(mu_s <@\textcolor{BrickRed}{+ 2}@>, sigma_s)
  b ~ normal(s, sigma_b <@\textcolor{BrickRed}{/ 100}@>)
  logit_o = s * lambda_so + b * lambda_bo
  o ~ bernoulli(1/(1+exp(-logit_o)))
end
\end{lstlisting}
\vspace{-2mm}
\caption{\enquote{Belief and skill matter} with assessment.}
    \end{subfigure}
    ~
    \begin{subfigure}[t]{0.49\textwidth}
        \label{fig:assessed_code_no_edge}
\begin{lstlisting}[frame=tb, framesep=3pt, xrightmargin=4pt, xleftmargin=0pt, style=Gen, numbers=none]
quote
  s ~ normal(mu_s <@\textcolor{BrickRed}{+ 2}@>, sigma_s)
  b ~ normal(s, sigma_b <@\textcolor{BrickRed}{/ 100}@>)
  logit_o = s * lambda_so
  o ~ bernoulli(1/(1+exp(-logit_o)))
end
\end{lstlisting}
\vspace{-2mm}
        \caption{\enquote{Only skill matters} with assessment.}
    \end{subfigure}
    
    \caption{Various interventions expressed as modifications of MiniStan source code.}
    \label{fig:interventions}
\end{figure*}

\section{Priors on Causal Models}
\label{sec:priors}

\begin{figure*}[t!]
    \centering
    \begin{subfigure}[t]{\textwidth}
        \centering
        \includegraphics[height=1.8in]{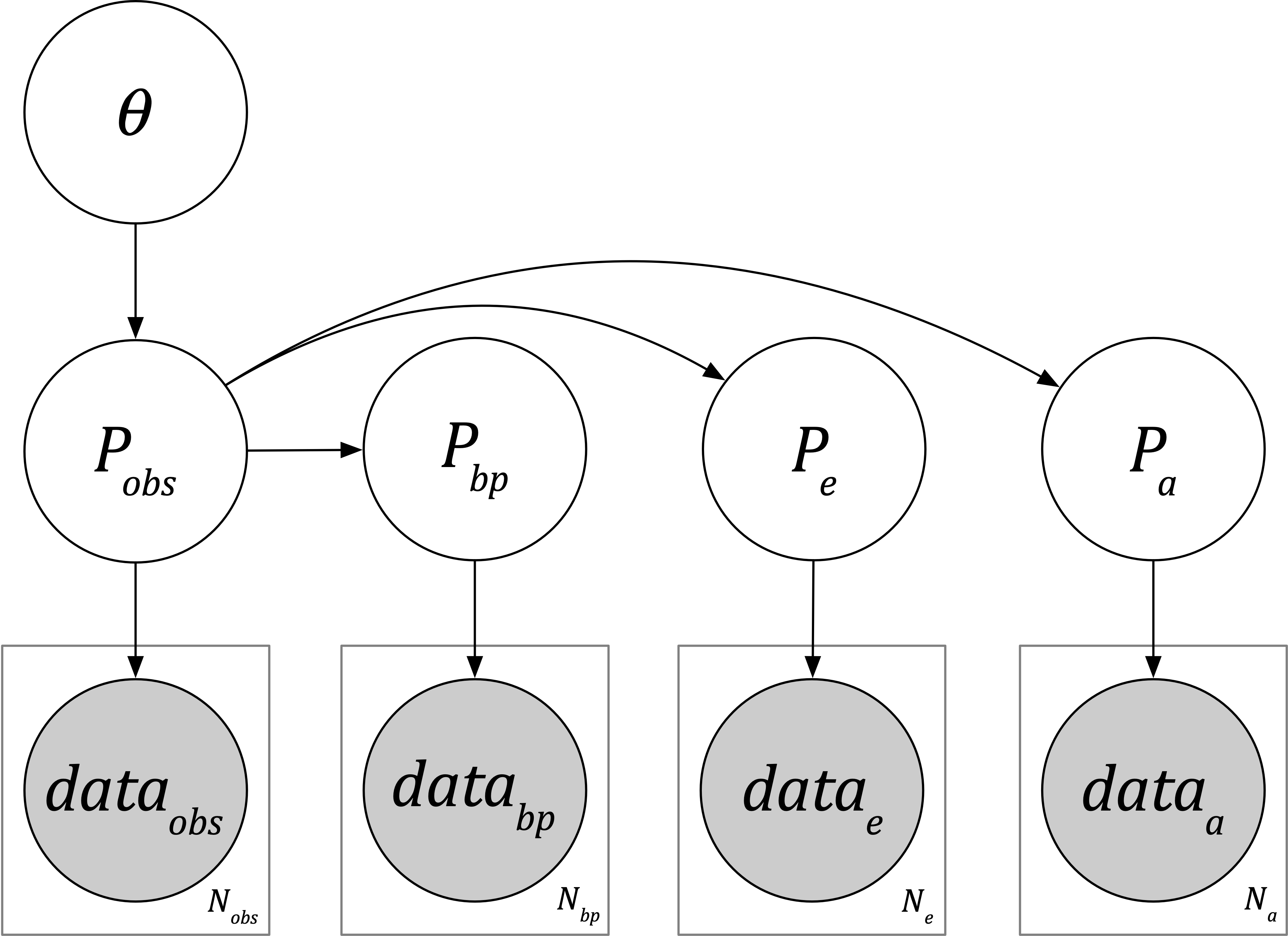}
    \end{subfigure}%
    \caption{Graphical meta-model for the Bayesian synthesis approach to causal structure and parameter learning. A set of global parameters $\theta$ determine the source code of the observational causal program $P_{obs}$, which is modified via code-editing intervention functions to induce experimental causal programs for the belief-pill ($P_{bp})$ encouragement design ($P_{e}$), and the assessment ($P_{a}$) interventions. The code for each program is run through an interpreter, which generates (observational or experimental) data. The likelihoods of the various kinds of data under the different interpreted programs can be used to infer the posterior distribution over $\theta$, and therefore over the observational causal program $P_{obs}$. 
    }
\end{figure*}
To compute the posterior distribution over the two candidate causal models, we first specify a prior distribution over a set of global latent variables. One of these variables, $edge$, determines whether $Belief$ influences $Outcome$.

\begin{align*}
 \mu_s &\sim Normal(0, 1) & \sigma_s &\sim Uniform(0, 1) & \sigma_b &\sim Uniform(0, 1) \\
 \lambda_{so} &\sim Uniform(0, 1) & \lambda_{bo} &\sim Uniform(0, 1) & edge &\sim Bernoulli(0.5)
\end{align*}
% \begin{align*}
%  Skill_i &\sim Normal(\mu_s, \sigma_s)\\ 
%  Belief_i &\sim Normal(Skill_i, \sigma_b)\\ 
%  OutcomeLogit_i &= \lambda_{so}~Skill_i + \lambda_{bo}~\mathbb{I}[edge]~Belief_i\\ Outcome &\sim Bernoulli(\frac{1}{1 + exp(-OutcomeLogit_i)})
% \end{align*}

\begin{figure*}[h!]
    \begin{subfigure}[t]{0.59\textwidth}
        \label{fig:gen_causal_model}
    \begin{lstlisting}[frame=tb, framesep=2pt, xrightmargin=4pt, xleftmargin=4pt, style=Gen]
@gen function generate_causal_model()
  mu_s = @trace(normal(0, 1), :mu_s)
  sigma_s = @trace(uniform(0, 1), :sigma_s)
  sigma_b = @trace(uniform(0, 1), :sigma_b)
  lambda_so = @trace(uniform(0, 1), :so_weight)
  lambda_bo = @trace(uniform(0, 1), :bo_weight)
  edge = @trace(bernoulli(0.5), :edge)
  
  if edge
    logit_o_expr = quote s * $so_weight + b * $bo_weight end
  else
    logit_o_expr = quote s * $so_weight end
  end
  
  causal_model = quote
    s ~ normal($mu_s, $sigma_s)
    b ~ normal(s, $sigma_b)
    logit_o = $logit_o_expr
    o ~ bernoulli(1/(1+exp(-logit_o)))
  end
  return causal_model
end
\end{lstlisting}
\vspace{-3mm}
\caption{}
    % \caption{Generative process for synthesizing causal models from the prior. Unlike the domain-specific language in which causal models are expressed (lines 13-18), Gen is a universal probabilistic language (enabling the \texttt{if} statement on line 8), and comes with a general-purpose inference engine.}
\end{subfigure}
\begin{subfigure}[t]{0.4\textwidth}
\begin{lstlisting}[frame=tb, framesep=2pt, xrightmargin=4pt, xleftmargin=4pt, style=Gen, numbers=none]
quote 
  s ~ normal(0.237, 0.449)
  b ~ normal(s, 0.913)
  logit_o = s * 0.137 + b * 0.852
  o ~ bernoulli(1/(1 + exp(-logit_o)))
end 
\end{lstlisting}
\begin{lstlisting}[frame=tb, framesep=2pt, xrightmargin=4pt, xleftmargin=4pt, style=Gen, numbers=none]
quote 
  s ~ normal(-0.592, 0.302)
  b ~ normal(s, 0.724)
  logit_o = s * 0.503 + b * 0.491
  o ~ bernoulli(1/(1 + exp(-logit_o)))
end 
\end{lstlisting}
\begin{lstlisting}[frame=tb, framesep=2pt, xrightmargin=4pt, xleftmargin=4pt, style=Gen, numbers=none]
quote 
  s ~ normal(1.892, 0.108)
  b ~ normal(s, 0.301)
  logit_o = s * 0.542
  o ~ bernoulli(1/(1 + exp(-logit_o)))
end 
\end{lstlisting}
\vspace{-3mm}
\caption{}
% \caption{Three example causal model programs sampled from the prior defined by \texttt{generate\_causal\_model}.}
\end{subfigure}

\begin{subfigure}[t]{\textwidth}
\begin{lstlisting}[frame=tb, framesep=2pt, xrightmargin=4pt, xleftmargin=4pt, style=Gen]
@gen function generate_data(NObs, NBeliefPill, NEncouragement, NAssessment)
  observational_model = @trace(generate_causal_model())
  belief_pill_model = applyDoIntervention(observational_model, :b, 5)
  encouragement_model = applyShiftIntervention(observational_model, :b, 3)
  assessment_model = applyVarianceScalingIntervention(applyShiftIntervention(observational_model, :s, 2),
                                                      :b, 1/100)
  
  observational_data = @trace(interpretMiniStan(observational_model, n_runs=NObs), :obs)
  belief_pill_data = @trace(interpretMiniStan(belief_pill_model, n_runs=NBeliefPill), :belief_pill)
  encouragement_data = @trace(interpretMiniStan(encouragement_model, n_runs=NEncouragement), :encouragement)
  assessment_data = @trace(interpretMiniStan(assessment_model, n_runs=NAssessment, :assessment)
end
\end{lstlisting}
\caption{}
\end{subfigure}

\caption{Gen implementation of causal inference via Bayesian synthesis. The \texttt{generate\_causal\_model} Gen program (a) encodes a prior distribution over MiniStan models; (b) shows three samples from this prior. The \texttt{generate\_data} Gen program (c) encodes the likelihood: it samples a possible causal model from the prior (line 2), modifies it to obtain MiniStan code representing experimental conditions (lines 3-6), then simulates observational and experimental data by running the MiniStan programs (lines 8-11). The interpreter is itself a Gen probabilistic program.}
\label{fig:gen_prior}
\end{figure*}

In the Bayesian synthesis framework, a prior distribution over causal models is a stochastic procedure generating programs in a domain specific language (Figure~\ref{fig:gen_prior}). The grammar for our simple domain specific language, MiniStan, is presented in Figure~\ref{fig:grammar}.

\section{Likelihoods for Experiments}

To incorporate experimental evidence of various forms, the Bayesian synthesis approach requires an intervention library which consists of a set of code-editing functions that modify causal model programs in the domain specific language. For the conceptual example, our intervention library contains three interventions: (i) an atomic intervention, which applies the do-operator; (ii) a shift intervention, which changes the mean of a distribution by a fixed increment; and (iii) a variance-scaling intervention, which modifies the variance of a random variable assumed to be drawn from a normal distribution. In principle, an intervention library could contain arbitrary rules for modifying causal model source code, including changing the underlying distribution for a random variable or adding variables (latent or observed) that didn't exist in the observational model.

These interventions can be freely composed to represent a diverse set of experimental scenarios. We demonstrate this compositionality in the \enquote{assessment} experiment, which is composed of a shift intervention (a student's skill may improve if she has to take a test) and a variance-scaling intervention (a student's belief in her ability has less noise after taking a test).

When interpreted, a causal program in MiniStan represents a likelihood function over observational data. To compute the likelihood of experimental data, we simply modify the causal program using the intervention library before subsequently interpretting the modified program.

\begin{figure*}[t!]
    \begin{subfigure}[t]{1\textwidth}
\begin{lstlisting}[frame=tb, framesep=2pt, xrightmargin=4pt, xleftmargin=4pt, style=Gen]
function applyDoIntervention(program, var, newValue)
  walk(program) do expr
    <@\textcolor{blue}{@match}@> expr begin
      :($x = $val)  && if x == var end => :($var = $newValue)
      :($x ~ $dist) && if x == var end => :($var = $newValue)
      _ => expr
    end
  end
end
\end{lstlisting}

\begin{lstlisting}[frame=tb, framesep=2pt, xrightmargin=4pt, xleftmargin=4pt, style=Gen]
function applyShiftIntervention(program, var, shiftValue)
  walk(program) do expr
    <@\textcolor{blue}{@match}@> expr begin
      :($x ~ normal($mean, $std)) && if x == var end => :($x ~ normal($mean + $shiftValue, $std))
      :($x ~ uniform($a, $b)) && if x == var end => :($x ~ uniform($a + $shiftValue, $b + $shiftValue))
      :($x = $value) && if x == var end => :($x = $value + $shiftValue)
      _ => expr
    end
  end
end
\end{lstlisting}
\end{subfigure}
%     ~
%     \begin{subfigure}[t]{0.3\textwidth}
%         \begin{lstlisting}[frame=tb, framesep=2pt, xrightmargin=20pt, xleftmargin=4pt, style=Gen]
% quote 
%     s ~ normal(0.237, 0.449)
%     b = 1
%     logit_o = s * 0.137 + b * 0.852
%     o ~ bernoulli(1/(1 + exp(-logit_o)))
% end
%         \end{lstlisting}
%         \begin{lstlisting}[frame=tb, framesep=2pt, xrightmargin=20pt, xleftmargin=4pt, style=Gen]
% quote 
%     s ~ normal(0.237, 0.449)
%     b = normal(s, 0.913)
%     logit_o = 1
%     o ~ bernoulli(1/(1 + exp(-logit_o)))
% end
%         \end{lstlisting}
%     \caption{Example applications of $do(b) = 1$ and $do(logit_o) = 1$.}
%     \end{subfigure}
\label{fig:do_intervention}
\caption{Julia implementation of the atomic (``do'') intervention and the shift intervention. Rather than perform graph operations such as removing edges, an atomic intervention on a program walks the program's code and replaces any expression that assigns \texttt{var} with a new expression, implementing the intervention (\texttt{var = newValue}). The shift intervention walks the program's code and adds \texttt{shiftvalue} to the mean argument for the normal distribution, the lower and upper bound arguments for the uniform distribution, and the value of any deterministic assignment.}
\end{figure*}

\section{Inference}

We demonstrate the utility of this approach by performing approximate posterior inference over synthesized causal model programs from our conceptual example. In this example we: (i) generate a MiniStan program from the prior, (ii) generate a set of observational and experimental data from the interpreted MiniStan program, and (iii) perform approximate posterior inference over synthesized causal models using sequential Monte Carlo~\cite{doucet2000sequential} with Metropolis Hastings rejuvination. We generated ten individuals' skill, belief, and outcome for each of the four observational and experimental settings from a single causal model where $\mu_s = -0.013, \sigma_s = 0.776, \sigma_b = 0.646, \lambda_{so} = 0.734, \lambda_{bo} = 0.717,$ and $edge=True$.

\begin{figure*}[h!]
    \centering
    \begin{subfigure}[l]{0.48\textwidth}
        \centering
        \includegraphics[height=1.6in]{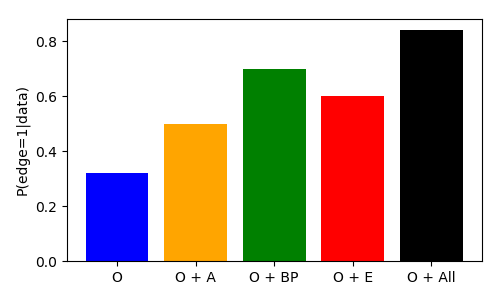}
    \end{subfigure}
    ~
    \begin{subfigure}[r]{0.48\textwidth}
        \centering
        \includegraphics[height=1.6in]{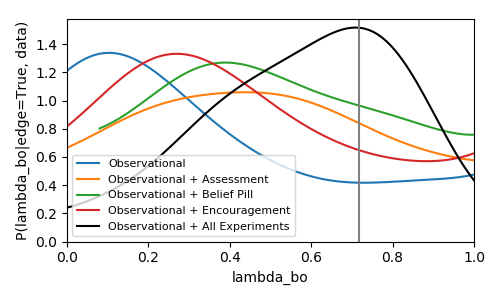}
    \end{subfigure}
    \caption{Posterior probability of the existence and strength of causal dependence between a student's belief and her subsequent outcome. The vertical gray line is the actual value for $\texttt{lambda\_{bo}}$}.
\end{figure*}

Using only observational data, the posterior probability of the edge variable is low. This may be because the data can be explained only by appealing to skill, and this simpler model could lead to a higher marginal probability than one which introduces a new parameter (\texttt{lambda\_bo}). (This phenomenon is sometimes called \enquote{Bayesian Ockham's Razor}.) However, as we incorporate additional experimental evidence the posterior probability of the edge increases. Similarly, the posterior distribution over $\lambda_{bo}$, the effect of belief on outcome, concentrates around the true value as we leverage experimental evidence. 

\section{Discussion}

The Bayesian synthesis approach we have outlined in this paper provides several advantages over alternative approaches to structure discovery and parameter estimation in causal modeling: (i) an explicit characterization of uncertainty over model structures; (ii) a principled way to model diverse interventions; and (iii) a formalization that can be re-used in diverse problems, with varying degrees of prior knowledge, without requiring practitioners to design custom inferences for each use case.

Although this example uses parametric causal models, it is conceptually straightforward to use Gaussian processes and/or Dirichlet process mixture models for the functional forms of causal relationships \cite{saad2019bayesian}. It may thus be fruitful to develop Bayesian variants of existing non-parametric techniques for causal inference \cite{imbens2004nonparametric, louizos2017causal}.

 The results reported here were obtained using vanilla sequential Monte Carlo over the joint space of model structure, parameters, and the latent variables in each observation or experiment. In order for this approach to scale to complex models, hierarchical priors over models, and large datasets, we expect more powerful techniques will be necessary. However, the Gen platform provides programmable inference constructs~\cite{cusumano-towner2019gen}, including hybrids of Hamiltonian Monte Carlo~\cite{duane1987hybrid} and Metropolis-Adjusted Langevin~\cite{roberts1996exponential} approaches with sequential Monte Carlo~\cite{doucet2000sequential}, that could potentially address some of these scaling challenges.

\section{Related Work}

Probabilistic programs are often used to represent causal processes~\cite{goodman2012church}. Some languages, such as Omega~\cite{tavares2019counterfactual}, make this causal interpretation explicit, including a semantics for interventional and counterfactual reasoning. It would be interesting to consider whether the framework we present here, which considers interventions to be arbitrary code-editing procedures, could also be usefully applied to counterfactual reasoning problems.

Incorporating experimental evidence for structure learning and parameter estimation can be thought of as the inner loop of an optimal experimental design procedure. Probabilistic programs have been used to automate this search over experiments~\cite{ouyang2016practical}, seeking to maximize the expected information gain over some query given new evidence. In that work, experiments are modeled as arguments to a probabilistic program. Our approach instead describes an experiment as a modification of MiniStan programs, enabling a clean abstraction between the specification of causal models (or distributions over causal models) and interventions that modify those models.

Improving methodology for combining observational and experimental evidence has far-reaching implications for a wide variety of scientific disciplines, and has received significant attention in the graph-based causal inference literature. For example, extensions of the do-calculus have been developed to incorporate experiments expressed as atomic interventions given a known causal graphical model structure~\cite{lee2019general}. Recent extensions of existing graph-based structure discovery algorithms have been made to incorporate atomic interventions~\cite{wang2017permutation} and imperfect interventions~\cite{yang2018characterizing}. Our work proposes characterizing imperfect interventions as code-editors acting on probabilistic programs; this representation enables us to perform posterior inference (with uncertainty estimates) over both structure and model parameters.

\section*{Acknowledgements} 

We thank Javier Burroni, Dan Garant, Zenna Tavares, and Reilly Grant for thoughtful discussion.

\bibliographystyle{acm}
\bibliography{ref}

\begin{thebibliography}{10}

\bibitem{aldrich1989autonomy}
{\sc Aldrich, J.}
\newblock Autonomy.
\newblock {\em Oxford Economic Papers 41}, 1 (1989), 15--34.

\bibitem{carpenter2017stan}
{\sc Carpenter, B., Gelman, A., Hoffman, M.~D., Lee, D., Goodrich, B.,
  Betancourt, M., Brubaker, M., Guo, J., Li, P., and Riddell, A.}
\newblock Stan: A probabilistic programming language.
\newblock {\em Journal of statistical software 76}, 1 (2017).

\bibitem{cusumano-towner2019gen}
{\sc Cusumano-Towner, M.~F., Saad, F.~A., Lew, A.~K., and Mansinghka, V.~K.}
\newblock Gen: A general-purpose probabilistic programming system with
  programmable inference.
\newblock In {\em Proceedings of the 40th ACM SIGPLAN Conference on Programming
  Language Design and Implementation\/} (New York, NY, USA, 2019), PLDI 2019,
  ACM, pp.~221--236.

\bibitem{doucet2000sequential}
{\sc Doucet, A., Godsill, S., and Andrieu, C.}
\newblock On sequential monte carlo sampling methods for bayesian filtering.
\newblock {\em Statistics and computing 10}, 3 (2000), 197--208.

\bibitem{duane1987hybrid}
{\sc Duane, S., Kennedy, A.~D., Pendleton, B.~J., and Roweth, D.}
\newblock Hybrid monte carlo.
\newblock {\em Physics letters B 195}, 2 (1987), 216--222.

\bibitem{eberhardt2007interventions}
{\sc Eberhardt, F., and Scheines, R.}
\newblock Interventions and causal inference.
\newblock {\em Philosophy of Science 74}, 5 (2007), 981--995.

\bibitem{friedman2000being}
{\sc Friedman, N., and Koller, D.}
\newblock Being bayesian about network structure.
\newblock In {\em Proceedings of the Sixteenth conference on Uncertainty in
  artificial intelligence\/} (2000), Morgan Kaufmann Publishers Inc.,
  pp.~201--210.

\bibitem{goodman2012church}
{\sc Goodman, N., Mansinghka, V., Roy, D.~M., Bonawitz, K., and Tenenbaum,
  J.~B.}
\newblock Church: a language for generative models.
\newblock {\em arXiv preprint arXiv:1206.3255\/} (2012).

\bibitem{griffiths2009theory}
{\sc Griffiths, T.~L., and Tenenbaum, J.~B.}
\newblock Theory-based causal induction.
\newblock {\em Psychological review 116}, 4 (2009), 661.

\bibitem{haavelmo1944probability}
{\sc Haavelmo, T.}
\newblock The probability approach in econometrics.
\newblock {\em Econometrica: Journal of the Econometric Society\/} (1944),
  iii--115.

\bibitem{heckerman1995learning}
{\sc Heckerman, D., Geiger, D., and Chickering, D.~M.}
\newblock Learning bayesian networks: The combination of knowledge and
  statistical data.
\newblock {\em Machine learning 20}, 3 (1995), 197--243.

\bibitem{imbens2004nonparametric}
{\sc Imbens, G.~W.}
\newblock Nonparametric estimation of average treatment effects under
  exogeneity: A review.
\newblock {\em Review of Economics and statistics 86}, 1 (2004), 4--29.

\bibitem{korb2004varieties}
{\sc Korb, K.~B., Hope, L.~R., Nicholson, A.~E., and Axnick, K.}
\newblock Varieties of causal intervention.
\newblock In {\em Pacific Rim International Conference on Artificial
  Intelligence\/} (2004), Springer, pp.~322--331.

\bibitem{lee2019general}
{\sc Lee, S., Correa, J.~D., and Bareinboim, E.}
\newblock General identifiability with arbitrary surrogate experiments.

\bibitem{louizos2017causal}
{\sc Louizos, C., Shalit, U., Mooij, J.~M., Sontag, D., Zemel, R., and Welling,
  M.}
\newblock Causal effect inference with deep latent-variable models.
\newblock In {\em Advances in Neural Information Processing Systems\/} (2017),
  pp.~6446--6456.

\bibitem{mansinghka2006structured}
{\sc Mansinghka, V., Kemp, C., Tenenbaum, J., and Griffiths, T.}
\newblock Structured priors for structure learning.
\newblock In {\em Proceedings of the Twenty-Second Conference on Uncertainty in
  Artificial Intelligence (UAI 2006)\/} (2006).

\bibitem{mccandless2009bayesian}
{\sc McCandless, L.~C., Gustafson, P., and Austin, P.~C.}
\newblock Bayesian propensity score analysis for observational data.
\newblock {\em Statistics in medicine 28}, 1 (2009), 94--112.

\bibitem{murphy2012machine}
{\sc Murphy, K.~P.}
\newblock {\em Machine learning: a probabilistic perspective}.
\newblock MIT Press, 2012.

\bibitem{ouyang2016practical}
{\sc Ouyang, L., Tessler, M.~H., Ly, D., and Goodman, N.}
\newblock Practical optimal experiment design with probabilistic programs.
\newblock {\em arXiv preprint arXiv:1608.05046\/} (2016).

\bibitem{pearl2000causality}
{\sc Pearl, J.}
\newblock {\em Causality: models, reasoning and inference}, vol.~29.
\newblock Springer, 2000.

\bibitem{roberts1996exponential}
{\sc Roberts, G.~O., Tweedie, R.~L., et~al.}
\newblock Exponential convergence of langevin distributions and their discrete
  approximations.
\newblock {\em Bernoulli 2}, 4 (1996), 341--363.

\bibitem{saad2019bayesian}
{\sc Saad, F.~A., Cusumano-Towner, M.~F., Schaechtle, U., Rinard, M.~C., and
  Mansinghka, V.~K.}
\newblock Bayesian synthesis of probabilistic programs for automatic data
  modeling.
\newblock {\em Proceedings of the ACM on Programming Languages 3}, POPL (2019),
  37.

\bibitem{sherman2019intervening}
{\sc Sherman, E., and Shpitser, I.}
\newblock Intervening on network ties.
\newblock In {\em Proceedings of the International Conference on Uncertainty in
  Artificial Intelligence\/} (2019).

\bibitem{tavares2019counterfactual}
{\sc Tavares, Z., Zhang, X., Koppel, J., and Lezama, A.~S.}
\newblock Soft constraints for inference with declarative knowledge.

\bibitem{wang2017permutation}
{\sc Wang, Y., Solus, L., Yang, K., and Uhler, C.}
\newblock Permutation-based causal inference algorithms with interventions.
\newblock In {\em Advances in Neural Information Processing Systems\/} (2017),
  pp.~5822--5831.

\bibitem{yang2018characterizing}
{\sc Yang, K.~D., Katcoff, A., and Uhler, C.}
\newblock Characterizing and learning equivalence classes of causal dags under
  interventions.
\newblock {\em arXiv preprint arXiv:1802.06310\/} (2018).

\end{thebibliography}

\end{document}